\documentclass[11pt]{article}

\usepackage[preprint]{acl}

\usepackage{times}
\usepackage{latexsym}
\usepackage[T1]{fontenc}
\usepackage[utf8]{inputenc}
\usepackage{microtype}
\usepackage{inconsolata}
\usepackage{graphicx}
\usepackage{placeins}
\usepackage{booktabs}
\usepackage{amsmath}
\usepackage{amssymb}
\usepackage[most]{tcolorbox}
\newtcblisting{promptbox}{
  listing only,
  colback=gray!7, colframe=black!50,
  arc=2mm, boxrule=0.4pt, breakable,
  left=2mm, right=2mm, top=1mm, bottom=1mm,
  listing options={basicstyle=\footnotesize\ttfamily,
                   breaklines=true, columns=fullflexible,
                   keepspaces=true, upquote=true}
}

\title{Multi-Specialist LLM Relay System for Competitive Programming}

\author{
  \textbf{Andrei Mikhailov\textsuperscript{1}},
  \textbf{Mikhail Burtsev\textsuperscript{2}},
  \textbf{Alsu Sagirova\textsuperscript{3,1}}
  \\
  \textsuperscript{1}MIRAI
  \\
  \textsuperscript{2}London Institute for Mathematical Sciences
  \\
  \textsuperscript{3}AXXX
  \\
  \small{
    \href{mailto:mikhailov.a@miriai.org}{\texttt{mikhailov.a@miriai.org}}
    \quad
    \href{mailto:mb@lims.ac.uk}{\texttt{mb@lims.ac.uk}}
    \quad
    \href{mailto:alsu.sagirova@axxx.tech}{\texttt{alsu.sagirova@axxx.tech}}
  }
}

\begin{document}
\maketitle

\begin{abstract}
Large Language Models excel at code generation, yet competitive programming exposes a persistent failure mode: existing multi-agent pipelines distribute work over generic planner, coder, and debugger roles and delegate the choice of algorithmic technique to the backbone alone. We present MARS (Multi-Agent Relay of Specialized LLMs), a prompt-only framework in which each agent is a topic specialist---dynamic programming, graphs, strings, geometry, and so on---grounded by retrieval-augmented generation over an algorithm-theory corpus. Given a problem, retrieval selects a small team of relevant specialists; a starter writes an initial C++17 solution, and each subsequent turn runs the candidate against public examples in a sandbox, lets the active specialist keep, repair, or hand off the draft, and forwards a structured packet to the next specialist. A single infrastructure-fixer pass normalizes boilerplate at the end. On the CodeContests test split with Gemma 4, MARS reaches $0.624 \pm 0.006$ pass rate at $2.3$ recorded pipeline stages per task ($+14.4$ percentage points over direct prompting), closing most of the gap to CodeSIM ($0.731$) at $3.3{\times}$ lower wall-clock cost and substantially smaller variance in per-task token spend. The source code is available on GitHub: 
\url{https://github.com/fckand/mars}.
\end{abstract}

\section{Introduction}
\label{sec:intro}
Multi-agent LLM systems recently became a popular solution for complex tasks, including software development, mathematical reasoning, and even scientific discovery \citep{guo2024largelanguagemodelbased, tran2025multiagentcollaborationmechanismssurvey, chen2025surveyllmbasedmultiagentsystem}. Such tasks often require specialized domain knowledge to succeed. For example, solving competitive programming problems requires a combination of theoretical algorithmic insights and problem-specific context, and advanced mathematical reasoning tasks require a combination of reasoning skills and a strong theoretical background. 

Existing approaches generally assume that LLMs' massive pre-training assures their wide-range expertise and rely on large proprietary pre-trained models as universal experts. This leads multi-agent systems to treat agent roles as generic abstractions rather than as carriers of real domain expertise. 

In this work, we propose a framework for a self-organized team of domain-specialized agents called MARS~--~Multi-Agent Relay of Specialized LLMs~--~a multi-agent framework featuring RAG-specialized agents for solving competitive programming problems.

Competitive programming has become a standard stress test for code-generating LLMs because the tasks require careful implementation and verification under sparse signal \citep{li2022competition,chen2023codet,islam2025codesim}. Moreover, problems often blend multiple theoretical areas to challenge algorithmic knowledge. Existing multi-agent approaches \citep{islam2024mapcoder, islam2025codesim, li2026solvitaenhancinglargelanguage} treat competitive programming problems as general code generation. They apply teams of planner, coder, and debugger agents that are generic with respect to the algorithmic content of the task, and topic competence is expected to emerge from the underlying LLMs. Content-agnostic pipelines provide no mechanism to supply the algorithmic expertise crucial for a correct solution. To address this gap, we propose MARS (Multi-Agent Relay of Specialized LLMs), a framework in which each agent is a domain expert specialized in a single algorithmic topic, grounded through retrieval-augmented generation. Given a problem, all available agents are asked two questions: whether the task matches their specialization and whether the agent can initialize the relay. Then, a small team of task-matched agents is formed. The initial agent generates the candidate solution (with iterative refinement based on public test execution results) and selects the next contributing agent from the team. The relay terminates when an agent judges the solution complete.

We propose MARS, a topic-aligned multi-agent relay in which each agent is a single-domain expert grounded by RAG over an algorithmic theory corpus, replacing the stage-aligned planner-coder-debugger decomposition. We make public-test execution an in-loop signal at every relay step: the same specialist sees its draft's report before keeping, repairing, or handing off. On CodeContests with Gemma 4, MARS reaches $0.624 \pm 0.006$ pass rate at $2.3$ recorded pipeline stages per task ($+14.4$ percentage points over direct prompting), closing most of the gap to CodeSIM ($0.731$) at $3.3{\times}$ lower wall-clock cost and ${\sim}7{\times}$ smaller standard deviation in per-task token spend.

\section{Related Work}
\label{sec:related}

Heterogeneous multi-agent LLM systems have been studied along the axes of backbone diversity \citep{ye2025xmasbuildingmultiagentsystems}, decentralized coordination without central orchestrators \citep{yang2025agentnet}, and dynamic teaming of capability-described agents drawn from a shared pool \citep{yun2026graphofagents, chen2025internetofagents}. A parallel line equips agents with persona or professional heterogeneity: inception-prompted role-playing \citep{li2023camelcommunicativeagentsmind}, medical specialties for clinical reasoning \citep{tang2024medagentslargelanguagemodels}, Thinker/Judge/Executor roles for mathematics \citep{lei2024macmutilizingmultiagentcondition}, and stacked heterogeneous LLM layers \citep{wang2024mixtureofagentsenhanceslargelanguage}. Across these systems heterogeneity is realized through personas, backbone diversity, evolving graph connections, or generic capability descriptions. None of them couples agent specialization to the topic structure of the task or grounds each specialist in a topical knowledge corpus, which is the gap MARS targets.

Several studies used retrieval mechanisms to improve code generation accuracy. REDCODER \citep{parvez-etal-2021-retrieval-augmented} retrieves relevant code or summaries from a database and supplies them to the generator; DocPrompting \citep{zhou2023docpromptinggeneratingcoderetrieving} retrieves library documentation in response to a natural-language intent; and RepoCoder \citep{zhang-etal-2023-repocoder} uses the task-supplied repo as a retrieval database for repository-level code completion; closer to deployment, \citet{wang2025copilottesting} retrieve context from an evolving codebase to synthesize tests and detect bugs. All of these retrieve over code, API documentation, or the codebase itself. MARS instead retrieves over distinct algorithmic-theory topics, one corpus slice per agent, which targets the algorithmic expertise a contest task needs rather than its implementation surface.

Once agents are heterogeneous the team is no longer fixed, and a growing body of work forms it at inference time: recruiting experts from task-conditional descriptions \citep{chen2023agentverse}, generating both agents and plan from the task specification on the fly \citep{chen2024autoagentsframeworkautomaticagent}, ranking candidates by an unsupervised importance score \citep{liu2023dynamic}, or optimizing node prompts together with inter-agent edges over a graph of LLM operations \citep{zhugegptswarm}. MARS instead organizes teams through self-reported topical competence: every specialist decides whether the problem falls within its expertise, and the matching specialists form the team.

Most progress in code generation has come from pairing a single strong base model with an outer loop that searches, verifies, or repairs its outputs: execution-based evaluation \citep{chen2021codex}, executed test cases \citep{chen2023codet}, execution-grounded debugging \citep{chen2023teachinglargelanguagemodels}, iterative self-feedback \citep{madaan2023selfrefine}, and verbal reinforcement from past trials \citep{shinn2023reflexionlanguageagentsverbal}; a complementary line aligns noisy crowd-sourced human feedback for RL-based code generation \citep{wong2025crlhf}, whereas MARS stays prompt-only and takes its feedback from deterministic public-test execution. Competitive programming is the principal stress test of this paradigm, because its problems combine a sparse correctness signal with deep algorithmic content: AlphaCode \citep{li2022competition} reached contest level only through enormous sampling with strong filtering, and later benchmarks report that even strong models solve a small fraction of olympiad problems, with the hardest tiers unsolved \citep{jain2024livecodebench, shi2024language}. 

Multi-agent approaches to code generation respond to this difficulty by decomposing the task into roles. AgentCoder \citep{huang2024agentcodermultiagentbasedcodegeneration} couples a programmer with a test designer and a test executor that iterate on each other's feedback; MapCoder \citep{islam2024mapcoder} chains retrieval, planning, coding, and debugging agents in a pipeline aimed at competitive problem solving; CodeSIM \citep{islam2025codesim} continues this line with simulation-driven planning and debugging; and Solvita \citep{li2026solvitaenhancinglargelanguage} develops a related multi-agent decomposition for the same setting. 
\begin{figure*}[htbp]
  \centering
\includegraphics[width=0.9\textwidth]{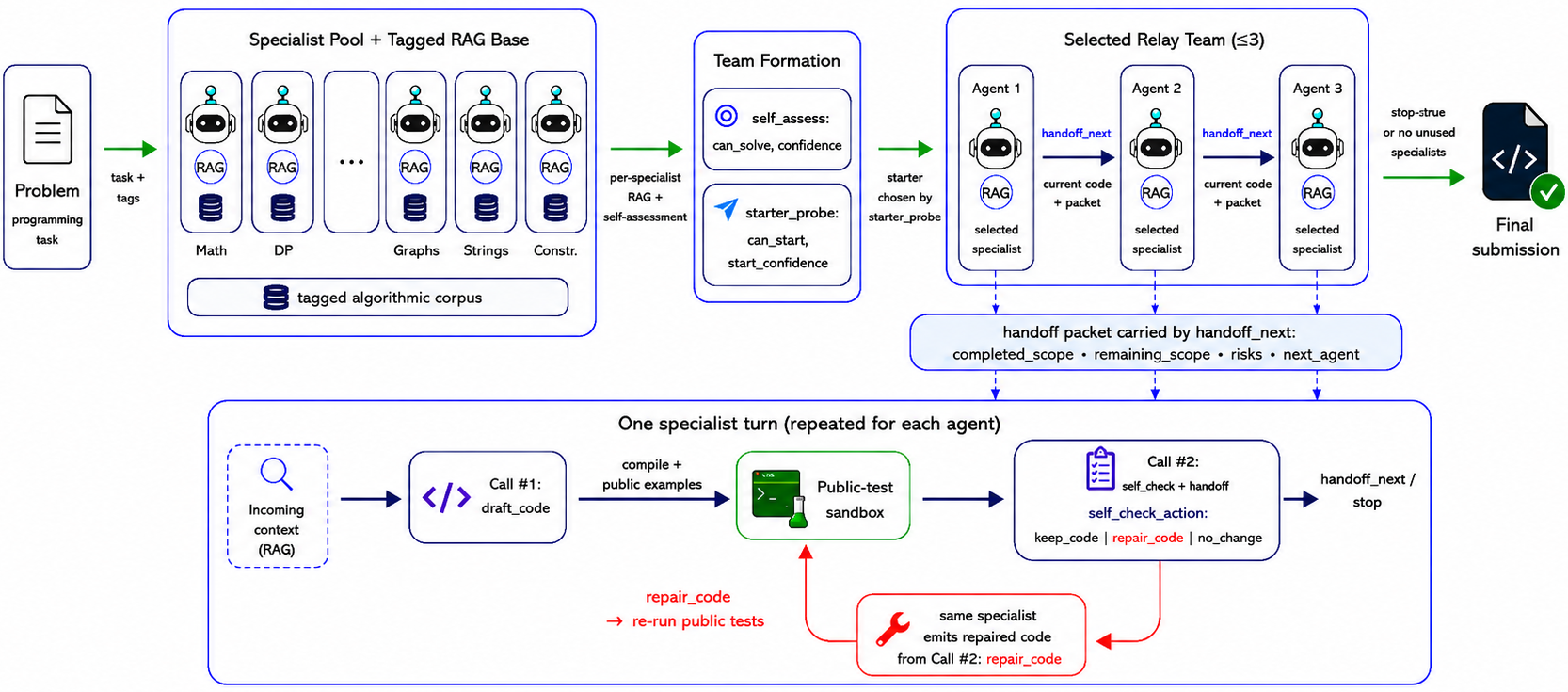}
  \caption{\textbf{MARS relay pipeline.} A task is routed from a pool of RAG-grounded topic specialists to a team of at most three agents. Each turn runs code generation, public-test execution, and self-check/handoff; repair code is rerun locally before the current code and relay packet move to the next specialist or final submission.}
  \label{fig:relay}
\end{figure*}

\section{Method}
\label{sec:method}
Figure~\ref{fig:relay} summarizes MARS. From a pool of eleven topic-specialized agents, a per-specialist self-assessment over a shared retrieval corpus shortlists a small team that relays a single C++17 program through self-checking handoffs. Full prompt templates and an end-to-end example are given in Appendices~\ref{app:prompts} and~\ref{app:example}; Appendix~\ref{app:pseudocode} gives pseudocode for MARS and for every baseline.

Each specialist is identified by a topic description and a tag set, and self-assesses against the shared cp-algorithms corpus filtered by its own tags. The assessment returns an in-scope flag, a relevance flag, and a confidence score. We shortlist up to three matches by these scores and pick the starter with a separate can-start probe.

Each turn issues two LLM calls. The first writes a draft from the current code, the assigned subtask, the starter contract, a compact summary of the previous relay state, and retrieved RAG context. The draft is executed against the public examples in ExecEval \citep{khan2023xcodeeval}. The second call sees the report and returns one of keep-code, repair-code, or no-change together with structured handoff fields. A repair candidate is rerun on the public examples and accepted only if it compiles and does not reduce the number of passing public tests relative to that turn's draft; otherwise the repair is rejected and the draft is restored. This deterministic local gate governs the keep/repair decision using observable execution signal rather than self-reported confidence: confidence scores enter only at team selection. The relay is budget-bounded to at most three unique specialists and eight steps, and stops on an explicit stop signal, when no unused selected specialist remains, at the step budget, or at a no-progress cutoff that reroutes at a streak of two and stops at three. After the relay the code is sanitized, and an infrastructure-fixer is invoked only when boilerplate-level failures (I/O wiring, includes, type widths) are detected.

We evaluate on 165 tasks from the CodeContests test split \citep{li2022competition}. The backbone is instruction-tuned Gemma 4\footnote{\url{https://huggingface.co/google/gemma-4-31B-it}}~\citep{gemma4team2026gemma4} with temperature $0.0$, top-$p$ $0.95$, and a $4096$ token budget. Retrieval uses the cp-algorithms corpus\footnote{\url{https://github.com/cp-algorithms/cp-algorithms}} encoded with Jina Embeddings~v2. 
All Table~\ref{tab:main} systems use Gemma~4, temperature $0.0$, and a $4096$-token budget. MARS, Parallel ensemble, Base relay, and CodeSIM use top-$p$ $0.95$; logged Direct and Single-RAG runs use $1.0$. Transfer runs follow their recorded model- and method-specific settings (Appendix~\ref{app:baselines}). Direct uses one call; Single-RAG uses the top retrieved specialist; Parallel ensemble merges specialist candidates; Base relay omits public-test self-check, subtask tracking, and the infrastructure-fixer. We adapt CodeSIM's open-source harness to the same $165$ tasks. Its published GPT-4 result uses a $156$-task subset \citep{islam2024mapcoder,islam2025codesim} and is not directly comparable.

\section{Results}
\label{sec:results}
\subsection{Main results}
\begin{figure}[!ht]
  \centering
\includegraphics[width=\linewidth]{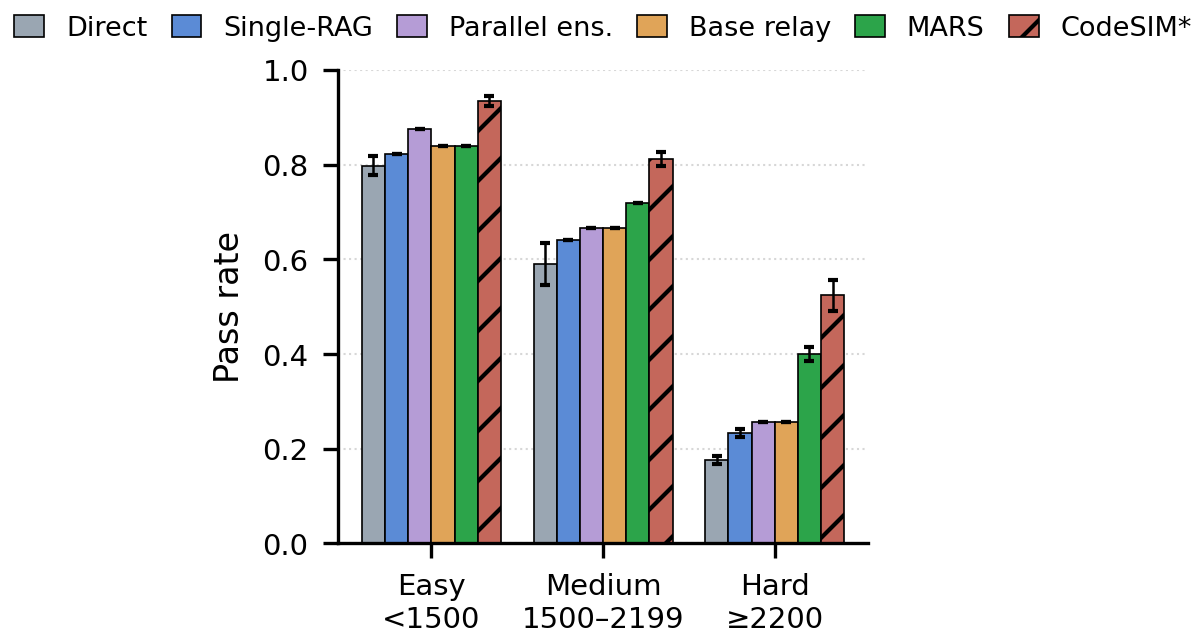}
  \caption{\textbf{Average pass rate by Codeforces difficulty tier.} MARS dominates prompt-only baselines on Medium and more than doubles Direct on Hard; CodeSIM$^{\ast}$ leads on every tier. Tier sizes: Easy $n{=}56$, Medium $n{=}39$, Hard $n{=}70$.}
  \label{fig:difficulty}
\end{figure}

Table~\ref{tab:main} summarizes the main results. MARS reaches $0.624 \pm 0.006$ at $2.3$ recorded pipeline stages per task, improving over Direct ($+0.144$), Single-RAG ($+0.095$), and the Parallel ensemble baseline ($+0.060$). CodeSIM reaches $0.731 \pm 0.009$; our method narrows this gap while using a simpler protocol with execution feedback at every specialist turn.
\begin{table}[htbp]
  \centering
  \footnotesize
  \setlength{\tabcolsep}{1pt}
  \newcommand{\pmcell}[2]{$#1\,{\scriptstyle\pm\,#2}$}
  \begin{tabular*}{\linewidth}{@{\extracolsep{\fill}}lcccc@{}}
    \toprule
    Method & Pass rate & Sec & Tokens & Calls \\
    \midrule
    Direct        & \pmcell{0.48}{0.02} & \pmcell{34.9}{49.6} & \pmcell{1.8}{1.2} & \pmcell{1.0}{0.0} \\
    Single-RAG    & \pmcell{0.53}{0.01} & \pmcell{59.8}{21.2} & \pmcell{25.0}{3.2} & \pmcell{12.0}{0.0} \\
    Parallel ens. & \pmcell{0.56}{0.00} & \pmcell{360.9}{172.9} & \pmcell{29.8}{5.1} & \pmcell{17.6}{0.8} \\
    Base relay    & \pmcell{0.55}{0.00} & \pmcell{191.6}{91.5} & \pmcell{34.1}{10.2} & \pmcell{17.1}{1.7} \\
    \textbf{MARS} & $\mathbf{0.62}\,{\scriptstyle\boldsymbol{\pm}\,\mathbf{0.01}}$ & \pmcell{244.3}{154.4} & \pmcell{40.3}{8.1} & \pmcell{16.6}{1.3} \\
    \midrule
    CodeSIM$^{\ast}$ & \pmcell{0.73}{0.01} & \pmcell{817.5}{1358.5} & \pmcell{32.2}{54.8} & \pmcell{10.2}{14.2} \\
    \bottomrule
  \end{tabular*}
  \caption{\textbf{Main results.} Pass rate is solved-task fraction. Time in sec, tokens in thousands and number of calls are per-task. All values averaged across 3 runs. CodeSIM$^{\ast}$ is our rerun on Gemma4.}
  \label{tab:main}
\end{table}
Among prompt-only baselines, Single-RAG plateaus at $0.529$ because a single specialist with no test signal cannot recover from algorithmic missteps, and Parallel ensemble pays $360.9\,$s per task on isolated candidates reconciled only at merge time. MARS adds in-step public-test feedback and opens a $+0.14$ gap on Hard tasks, where prompt-only baselines hover near $0.18$--$0.26$ (Figure~\ref{fig:difficulty}). CodeSIM still leads on every tier via up to $45$ debug iterations per Hard task, but MARS closes most of the gap at $3.3{\times}$ lower wall-clock cost by routing to a topic specialist instead of re-planning a generic solution.
\begin{figure}[htbp]
  \centering
\includegraphics[width=\linewidth]{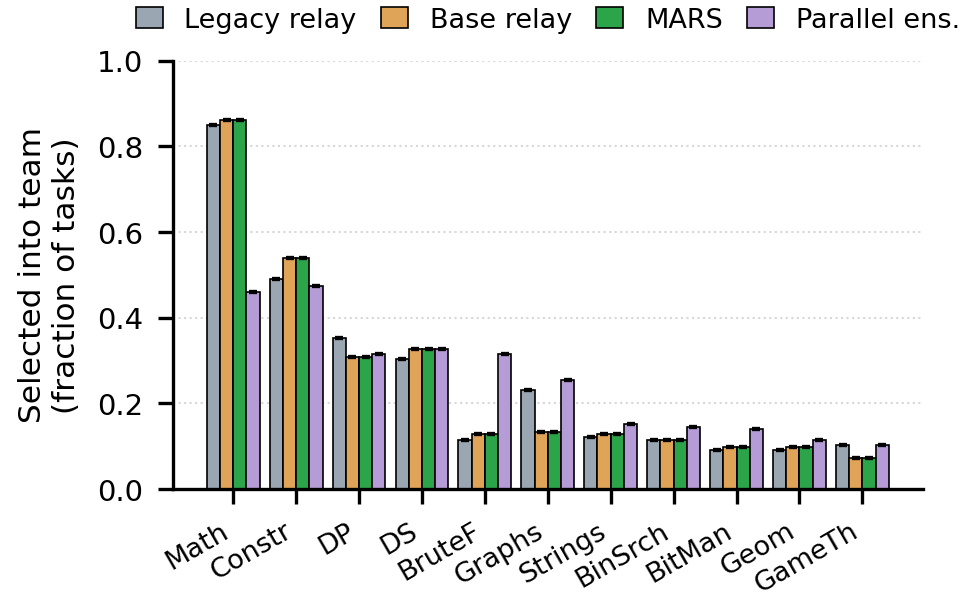}
  \caption{\textbf{Team-selection frequency per specialist.} Fraction of tasks on which each specialist joins the team, averaged over runs. Base relay and MARS share a deterministic assessor and coincide; the Parallel ensemble uses an earlier, less selective prompt and spreads selections more widely, which does not translate into accuracy.}
  \label{fig:agent_selection}
\end{figure}
Routing concentrates on Mathematics, Constructive Algorithms, Data Structures, and Dynamic Programming; rarer specialists fire only on tag-matched tasks (Figure~\ref{fig:agent_selection}). Base relay and MARS share the same deterministic assessor and therefore coincide in team distributions. Their difference is downstream of selection and reflects the combined update to public-test self-check, subtask tracking, and final infrastructure handling. 
Parallel ensemble's earlier assessor admits more borderline specialists (Brute Force, Graphs), but broader selection alone does not improve accuracy without execution feedback. Base relay and MARS share the subtask graph and starter contract; the broader MARS protocol adds public-test self-check and final infrastructure handling and reaches $0.624$ rather than $0.552$, although this comparison does not isolate the contribution of each change. Scores remain near ceiling on Easy ($0.80$--$0.93$), while MARS's advantage over Direct widens on Medium ($0.72$ vs.\ $0.59$) and Hard ($0.40$ vs.\ $0.18$; Figure~\ref{fig:difficulty}).

\subsection{Other backbones and target languages}
\label{sec:generalization}
Table~\ref{tab:backbones} keeps tasks, methods, and final evaluation fixed while following each run's recorded API settings. MARS has the highest pass rate of the three methods on every backbone, beating Single-RAG by $9.5$ points on Gemma~4, $3.3$ on Qwen3.5-27B, and $13.9$ on GPT-5.4-mini. The ordering Direct $<$ Single-RAG $<$ MARS therefore persists across backbones. On Python, MARS reaches $0.622 \pm 0.015$ against Direct's $0.485 \pm 0.000$ ($+13.7$ points); both match their C++17 counterparts within uncertainty, although Python is slower ($307.6$ vs.\ $244.3$\,s).

PairCoder \citep{zhang2024paircoder} is a Navigator/Driver MAS with multi-plan search. Our adapter preserves its role flow and prompts but replaces the dataset, model API, and execution boundaries. It reaches $0.705 \pm 0.009$, $8.3$ points above MARS at $1.4{\times}$ the wall-clock cost, making it a strong, heavier-search competitor. Its plan clustering uses a proprietary embedding model; the matched Python MARS run uses open-weight components throughout.

\begin{table}[htbp]
  \centering
  \footnotesize
  \setlength{\tabcolsep}{0pt}
  \newcommand{\pmcell}[2]{$#1\,{\scriptstyle\pm#2}$}
  \begin{tabular*}{\linewidth}{@{\extracolsep{\fill}}lcccc@{}}
    \toprule
    Method & Pass rate & Sec & Tokens & Calls \\
    \midrule
    \multicolumn{5}{@{}l}{\emph{Qwen3.5-27B, C++17}} \\
    Direct     & \pmcell{0.192}{0.018} & \pmcell{27.2}{74.6} & \pmcell{1.7}{2.2} & \pmcell{1.0}{0.0} \\
    Single-RAG & \pmcell{0.264}{0.004} & \pmcell{71.0}{57.1} & \pmcell{26.5}{4.0} & \pmcell{12.0}{0.0} \\
    \textbf{MARS} & \pmcell{\mathbf{0.297}}{\mathbf{0.012}} & \pmcell{109.7}{99.7} & \pmcell{40.5}{18.7} & \pmcell{15.3}{3.3} \\
    \midrule
    \multicolumn{5}{@{}l}{\emph{GPT-5.4-mini, C++17}} \\
    Direct     & \pmcell{0.149}{0.019} & \pmcell{3.4}{2.3} & \pmcell{1.2}{0.5} & \pmcell{1.0}{0.0} \\
    Single-RAG & \pmcell{0.364}{0.024} & \pmcell{49.7}{26.3} & \pmcell{24.7}{3.2} & \pmcell{12.0}{0.1} \\
    \textbf{MARS} & \pmcell{\mathbf{0.503}}{\mathbf{0.043}} & \pmcell{92.2}{61.1} & \pmcell{36.8}{7.4} & \pmcell{15.8}{1.8} \\
    \midrule
    \multicolumn{5}{@{}l}{\emph{Gemma 4, Python (PyPy 3)}} \\
    Direct     & \pmcell{0.485}{0.000} & \pmcell{70.0}{137.8} & \pmcell{1.6}{1.0} & \pmcell{1.0}{0.0} \\
    MARS       & \pmcell{0.622}{0.015} & \pmcell{307.6}{215.7} & \pmcell{42.9}{9.1} & \pmcell{17.1}{1.4} \\
    \textbf{PairCoder} & \pmcell{\mathbf{0.705}}{\mathbf{0.009}} & \pmcell{426.7}{451.7} & \pmcell{27.6}{20.9} & \pmcell{7.6}{5.4} \\
    \bottomrule
  \end{tabular*}
  \caption{\textbf{Backbone and language transfer.} The same $165$ tasks and final harness as Table~\ref{tab:main}, using recorded model- and method-specific decoding. Time is in seconds, tokens in thousands, and calls per task; PairCoder includes embedding traffic.}
  \label{tab:backbones}
\end{table}

\subsection{Ablations and protocol variants}
\label{sec:ablation}
Table~\ref{tab:ablation} combines one RAG ablation with broader protocol variants. Removing RAG alone costs $2.0$ points. Generalists without RAG are $0.9$ points below MARS within one standard deviation, but take $31\%$ longer ($319.8 \pm 245.1$ vs.\ $244.3 \pm 154.4$\,s) and more calls ($17.3 \pm 1.3$ vs.\ $16.6 \pm 1.3$). Because retrieval also changes, this row does not isolate specialization. Base relay and Parallel manager alter several post-selection components and trail MARS by $7.2$ and $6.0$ points; the full relay remains strongest among our configurations.

\begin{table}[htbp]
  \centering
  \footnotesize
  \newcommand{\pmcell}[2]{$#1\,{\scriptstyle\pm#2}$}
  \begin{tabular*}{\linewidth}{@{\extracolsep{\fill}}lcc@{}}
    \toprule
    Configuration & Pass rate & $\Delta$ \\
    \midrule
    MARS (full)                     & \pmcell{0.624}{0.006} & --- \\
    \quad w/o RAG grounding         & \pmcell{0.604}{0.007} & $-0.020$ \\
    \quad Generalists, no RAG       & \pmcell{0.615}{0.013} & $-0.009$ \\
    \quad Earlier Base relay        & \pmcell{0.552}{0.000} & $-0.072$ \\
    \quad Parallel manager          & \pmcell{0.564}{0.000} & $-0.060$ \\
    \bottomrule
  \end{tabular*}
  \caption{\textbf{Ablations and protocol variants} on Gemma 4. The generalist variant also disables RAG; the last two rows are the broader Base relay and Parallel ensemble comparisons from Table~\ref{tab:main}.}
  \label{tab:ablation}
\end{table}

\subsection{Relay behaviour and failure modes}
\label{sec:behaviour}
Teams contain one, two, or three agents on $1.8\%$, $15.8\%$, and $82.4\%$ of task-runs. An average of $1.35$ specialists change the code; the reported $2.3$ recorded stages also count the final sanitizer/fixer record and therefore measure pipeline-history depth, not specialist turns.

The next specialist receives shared code and a compact relay summary, not the raw public-test report. The gate compares a repair only with its same-turn draft; a later specialist may replace that code. It reverted $4.4 \pm 0.9\%$ of $697$ self-check decisions (Appendix~\ref{app:gate}).

Multi-topic tasks ($88\%$) need no reconciliation because specialists edit one shared draft sequentially. Pass rate is $0.612 \pm 0.010$ on multi-topic and $0.719 \pm 0.030$ on single-topic tasks. The boilerplate-only fixer changed one task ($\approx 0.2\%$ of task-runs), so the headline $0.624$ is independent of it (Appendix~\ref{app:fixer}).

\section{Conclusion}
MARS is a prompt-only, topic-aligned MAS that beats single-agent and ensemble baselines by $6$--$14$ percentage points, with public-test feedback at every specialist step, and holds that advantage across three backbones and two languages. Heavier-search systems---CodeSIM and PairCoder in Python---still lead on pass rate.

\section*{Limitations}

The evaluation covers $165$ CodeContests tasks, three backbones, two languages, one corpus, and Codeforces tags. Python reuses the same corpus and index; further languages need their own prompts, extraction, sandbox, and infrastructure. Transfer beyond C++17 and Python remains untested.

The local gate rejects only same-turn public-test regressions; it misses hidden tests and comparisons between specialists. All generated code requires sandboxed execution.

CodeSIM is the only stage-aligned comparison, and PairCoder remains Python-only. Other baselines need method-specific ports (Appendix~\ref{app:baselines}).

\bibliography{references}

\appendix

\section{Prompt Templates}
\label{app:prompts}

MARS uses four prompt templates at runtime. Each specialist is first queried with a self-assessment prompt (Figure~\ref{fig:prompt_assess}) that gates inclusion in the team. The shortlisted specialists then run a first-agent probe (Figure~\ref{fig:prompt_starter}) that elects the starter. Each relay turn afterwards consists of a code-generation call (Figure~\ref{fig:prompt_codegen}) followed by an execution-aware self-check and handoff call (Figure~\ref{fig:prompt_handoff}). Placeholders in \texttt{\{braces\}} are filled by the harness from per-agent metadata, the current task, relay state, and retrieved RAG context; long in-prompt examples are abbreviated for space.

\begin{promptbox}
You are {agent_name}.
Your specialty: {agent_description}

Analyze if this programming problem matches
YOUR specific expertise.

CRITICAL: Be selective! Only say "can_solve":
true if the problem DIRECTLY relates to your
specialty.

REQUIRED FORMAT:
{"can_solve": true/false,
 "is_relevant_to_specialty": true/false,
 "confidence": 0.0-1.0,
 "reasoning": "explain how you can help"}

EXAMPLE:
- GraphTheoryAgent + "Find shortest path" ->
  {"can_solve": true,
   "is_relevant_to_specialty": true,
   "confidence": 0.88,
   "reasoning": "Dijkstra applies here"}

GUIDELINES:
- ONLY answer true if the problem is DIRECTLY
  in your domain.
- If a major part of the task is in your
  specialty, answer is_relevant_to_specialty
  = true even when other specialties are also
  needed.

PROGRAMMING PROBLEM:
{task}

Relevant documentation:
{rag_context}

STRICT OUTPUT RULES:
- Return ONLY a single valid JSON object.
- Do not output thinking steps or chain-of-
  thought.
\end{promptbox}
\captionof{figure}{Specialist self-assessment prompt; the output gates inclusion in the relay team.}
\label{fig:prompt_assess}

\begin{promptbox}
You are {agent_name}, specialized in
{agent_description}.

{rag_context}

TASK:
{task}

You are evaluating whether you should be the
FIRST agent to start solving this programming
task.

Your job is NOT to describe the full solution.
Your job is to decide whether the task should
START in your specialty, and if yes, to define
exactly one narrow first contribution that
belongs to you.

Respond with ONLY one valid JSON object:
{
  "can_start": true/false,
  "start_confidence": 0.0-1.0,
  "owned_subproblem": "<one narrow sub-problem
      that belongs to your specialty>",
  "what_would_you_do_first": "<one concrete
      first step you would personally own>",
  "starter_reasoning": "<concrete local
      reasoning for only that first step>",
  "out_of_scope": ["<what you would explicitly
      NOT solve in the first step>"]
}

Rules:
- Be strict. If the task should not START in
  your specialty, set can_start=false.
- Keep the first step narrow and specialty-
  specific; do not describe the full algorithm.
- Do NOT write code in this stage.
- Do NOT output chain-of-thought or markdown
  fences.
\end{promptbox}
\captionof{figure}{First-agent probe; sets the starter contract that is later passed to every relay step.}
\label{fig:prompt_starter}

\begin{promptbox}
You are {agent_name}, specialized in
{agent_description}.

You are in a relay loop. This call is
CODE GENERATION ONLY.

TASK:
{task}

ASSIGNED SUBTASK:
{assigned_subtask}

STARTER CONTRACT (FROM FIRST-AGENT DECISION):
{starter_contract}

CURRENT CODE ({code_length} chars):
```cpp
{current_code}
```

YOUR SELECTION-TIME REASONING:
{agent_task_reasoning}

PREVIOUS HANDOFF:
{previous_handoff}

SUBTASK GRAPH:
{subtask_graph}

GATE FEEDBACK FROM RECENT STEPS:
{gate_feedback}

NO-PROGRESS STREAK: {no_progress_streak}

RAG CONTEXT:
{rag_context}

Respond with ONLY one valid JSON object:
{
  "action": "write_code" | "no_change",
  "new_code": "<complete C++17 code or empty>",
  "stop": true/false,
  "confidence": 0.0-1.0
}

Rules:
- Use "write_code" only when you provide full
  runnable C++17 in new_code.
- If action="no_change", new_code must be empty.
- stop=true only when the relay can finish now.
- If NO-PROGRESS STREAK >= 2, prefer a concrete
  fix over repeated no_change.
- new_code must contain ONLY raw C++ code (no
  markdown fences, prose, or JSON fragments).
- Code must contain zero comments.
\end{promptbox}
\captionof{figure}{Relay step call~1 (code generation). The draft is then executed on public examples in the sandbox before call~2.}
\label{fig:prompt_codegen}

\begin{promptbox}
You are {agent_name}, specialized in
{agent_description}.

You are in a relay loop. This call is
SELF-CHECK + HANDOFF ONLY.

TASK:
{task}

ASSIGNED SUBTASK:
{assigned_subtask}

STARTER CONTRACT (FROM FIRST-AGENT DECISION):
{starter_contract}

DRAFT CODE AFTER YOUR FIRST CALL
({draft_code_length} chars):
```cpp
{draft_code}
```

PUBLIC TEST REPORT FOR THE DRAFT:
{public_test_report}

PREVIOUS HANDOFF:
{previous_handoff}

SUBTASK GRAPH:
{subtask_graph}

GATE FEEDBACK FROM RECENT STEPS:
{gate_feedback}

NO-PROGRESS STREAK: {no_progress_streak}

AVAILABLE UNUSED AGENTS:
{available_agents}

Respond with ONLY one valid JSON object:
{
  "self_check_action":
      "keep_code" | "repair_code" | "no_change",
  "new_code": "<complete C++17 code or empty>",
  "self_check_summary": "<what you checked
      or fixed>",
  "test_report_interpretation": "<short
      interpretation of the public-test report>",
  "stop": true/false,
  "next_agent": "<agent name or null>",
  "completed_scope": "<what your full step
      completed>",
  "remaining_scope": "<what is still missing>",
  "resolved_subtasks": ["<ids/titles resolved>"],
  "unresolved_subtasks": ["<ids/titles pending>"],
  "handoff_instruction": "<explicit instruction
      for the next agent>",
  "known_risks": "<main risks/uncertainties
      or 'none'>"
}

Rules:
- Treat the public-test report as a diagnostic
  signal, not absolute proof.
- If the draft is already better than any
  possible repair, set self_check_action=
  "keep_code" and new_code="".
- Use self_check_action="repair_code" only when
  you return full runnable C++17 in new_code.
- If stop=false, next_agent is REQUIRED and
  must be one name from AVAILABLE UNUSED AGENTS.
- If stop=true or no unused agents remain,
  next_agent must be null.
- Do not add testing code, asserts, debug
  prints, or hardcoded sample answers.
- For repair_code, new_code must contain ONLY
  raw C++ code (no markdown fences, prose, or
  JSON fragments).
\end{promptbox}
\captionof{figure}{Relay step call~2 (execution-aware self-check and handoff). Output drives the keep/repair/skip decision and selects the next specialist.}
\label{fig:prompt_handoff}

\section{Example Run}
\label{app:example}

\begin{figure*}[t]
  \centering
  \includegraphics[height=0.82\textheight,keepaspectratio]{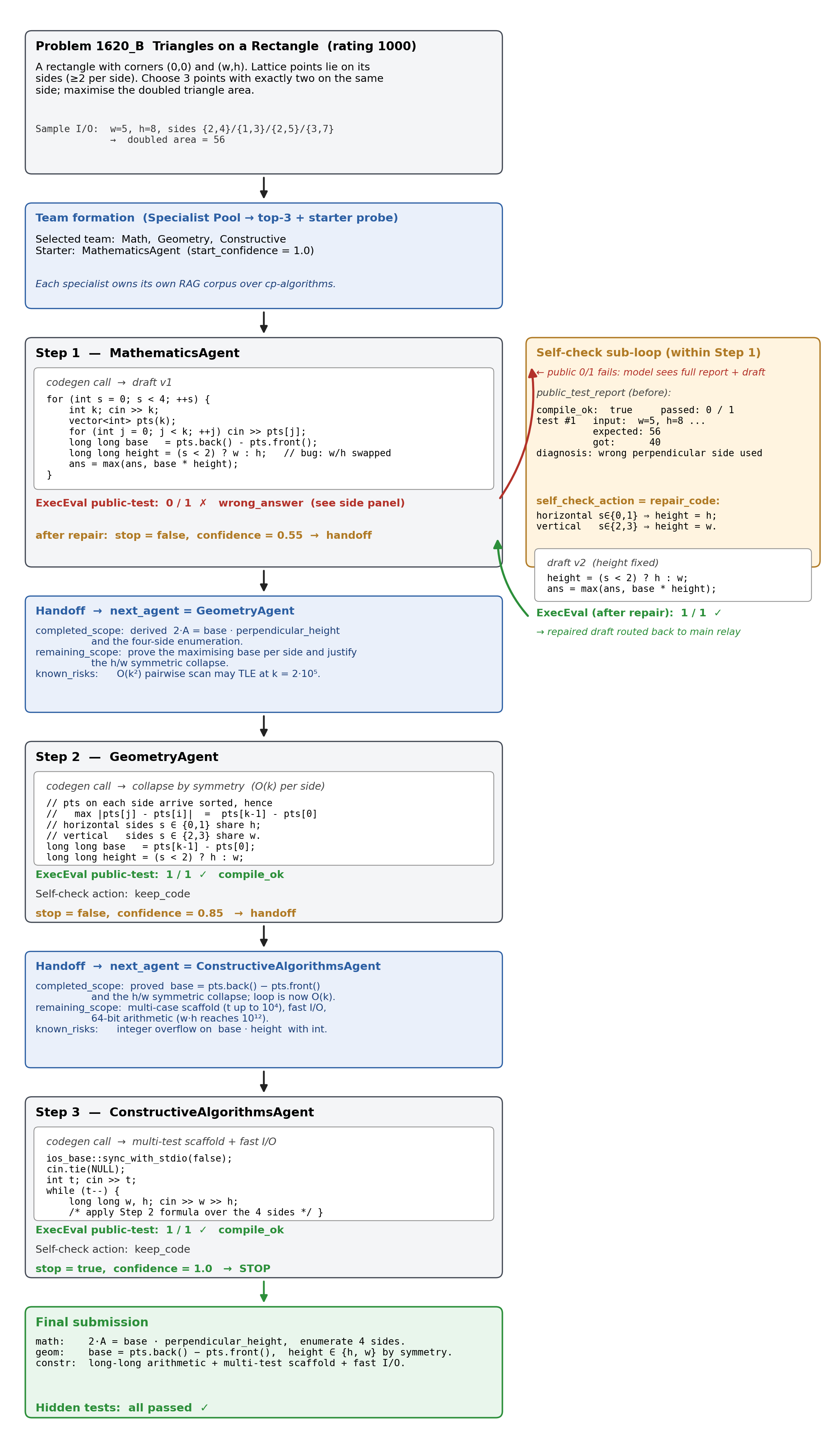}
  \caption{Example MARS trace for Codeforces 1620\_B with three contributing specialists. Step~1 includes a within-step self-check repair loop: the first draft fails the public sample, the same specialist consumes the test report and emits a corrected draft before handing off. Later steps show the specialist contribution, public-test outcome, and structured handoff to the next agent.}
  \label{fig:example}
\end{figure*}

Figure~\ref{fig:example} shows one successful MARS trace on Codeforces problem 1620\_B \emph{Triangles on a Rectangle}. The system selects three specialists: MathematicsAgent, GeometryAgent, and ConstructiveAlgorithmsAgent. MathematicsAgent derives the area formula and four-side enumeration; its first draft fails the public sample because the rectangle height and width are swapped for two sides, and the same specialist repairs the draft after seeing the public-test report. GeometryAgent then reduces each side to a single endpoint subtraction and clarifies the opposite-dimension height. ConstructiveAlgorithmsAgent completes the multi-test scaffold, fast I/O, and 64-bit arithmetic. The final program returned by the third relay step passes the hidden tests.

\section{Method Pseudocode}
\label{app:pseudocode}

For a side-by-side qualitative comparison of MARS with every baseline, Algorithms~1--6 give the control flow of each system in the form actually run in our harness. MARS forms a team of task-relevant specialists that sequentially edit one shared draft, with an intra-step refinement loop driven by public-test execution and a post-relay infrastructure check. Direct and Single-RAG are single-agent: the former prompts the backbone with the task description alone, the latter first elects one specialist from the pool. Parallel ensemble extends Single-RAG to several pre-selected specialists that generate independently, with a manager LLM aggregating the drafts. Base relay performs team formation and relay editing but no execution feedback. CodeSIM plans, simulates the plan, revises it, generates code, and debugs against the public tests until they pass or the attempt limit is reached.

\begin{promptbox}
Algorithm 1: MARS
Input: problem P, public examples S, pool A
1. For each specialist a in A:
       retrieve topic-matched context;
       self-assess capability, relevance, confidence.
2. Rank eligible specialists by tag overlap, RAG
   coverage, confidence, retrieval quality;
   select up to three; choose a starter.
3. Init shared draft C, handoff H, subtask state G.
4. While a step is permitted and a specialist is
   available:
       C_draft, H, next, stop <-
           active(P, C, G, H, retrieved context)
       R_before <- execute C_draft on S
       action, C_repair, H, stop <-
           self_check(C_draft, R_before)
       if action == repair_code:
           R_after <- execute C_repair on S
           if C_repair compiles and
              R_after.passed >= R_before.passed:
               C <- C_repair
           else:
               C <- C_draft
       else:
           C <- C_draft
       if stop, or no unused specialist remains,
          or no-progress cutoff fires: break
       active <- requested eligible specialist,
                 else fallback
5. Apply the infrastructure fixer only if
   boilerplate-level failure is detected; return C.
\end{promptbox}
\captionof{figure}{MARS: a selected team of task-relevant specialists relays solution updates with an intra-step self-refinement loop over public-test execution and a post-relay infrastructure check.}
\label{alg:mars}

\begin{promptbox}
Algorithm 2: Direct
Input: problem P
1. Prompt the LLM once with P and the requested
   C++17 output format.
2. C <- generated program.
3. Return C unchanged.
\end{promptbox}
\captionof{figure}{Direct: a single agent generates code from the task description.}
\label{alg:direct}

\begin{promptbox}
Algorithm 3: Single-RAG
Input: problem P, specialist pool A
1. Retrieve topic-matched context and self-assess
   each specialist in A.
2. a* <- highest-ranked eligible specialist.
3. C <- a* generates one program from P and its
   retrieved context.
4. Return C unchanged.
\end{promptbox}
\captionof{figure}{Single-RAG: the single most suitable specialist solves the task directly.}
\label{alg:singlerag}

\begin{promptbox}
Algorithm 4: Parallel ensemble
Input: problem P, specialist pool A
1. Retrieve context; select task-matched
   specialists from A.
2. Each selected specialist independently
   produces a plan and candidate code.
3. C <- manager LLM combines the candidates into
   one final program.
4. Return C.
\end{promptbox}
\captionof{figure}{Parallel ensemble: task-matched specialists solve independently in parallel and a general-purpose manager aggregates the candidates.}
\label{alg:parallel}

\begin{promptbox}
Algorithm 5: Base relay
Input: problem P, specialist pool A
1. For each specialist a in A:
       retrieve topic-matched context;
       self-assess capability, relevance,
       confidence.
2. Rank eligible specialists as in Algorithm 1;
   choose a starter.
3. Init shared draft C, handoff H, subtask state G.
4. While a step is permitted and an active
   specialist is available:
       C_draft, H, next, stop <- active(P, C, G, H)
       C <- C_draft
       if stop: break
       active <- requested eligible specialist,
                 else fallback
5. Return C.
\end{promptbox}
\captionof{figure}{Base relay: the team relays edits to a shared draft with no public-test execution inside the step.}
\label{alg:baserelay}

\begin{promptbox}
Algorithm 6: CodeSIM
Input: problem P, public examples S
for each plan attempt (up to the limit):
    plan <- PlanningAgent(P)
    simulation <- simulate plan on S
    if simulation requests revision:
        plan <- refine plan using the critique
    C <- CodingAgent(P, plan)
    passed, log <- execute C on S
    if passed: return C
    for each debugging attempt (up to the limit):
        C <- DebuggingAgent(P, plan, C, log)
        passed, log <- execute C on S
        if passed: return C
return the last generated C
\end{promptbox}
\captionof{figure}{CodeSIM: plan, simulate, revise, generate, then debug against the public tests.}
\label{alg:codesim}

\section{Relay Decision Statistics}
\label{app:gate}

Across the reported MARS runs the relay took $697$ self-check decisions. Of these, $38.4 \pm 1.0\%$ accepted a repair, $55.4 \pm 1.4\%$ kept the draft unchanged, $4.4 \pm 0.9\%$ proposed a repair that the gate rejected and reverted, and $1.7 \pm 0.8\%$ were compile failures the specialist did not repair. Repair acceptance requires non-regression against the same-turn draft on the public tests. This guarantee does not extend to a new draft written by the next specialist; over-inclusive selection is instead bounded by the shared-code workflow and the no-progress cutoff.

\section{Infrastructure-Fixer Statistics}
\label{app:fixer}

The infrastructure-fixer runs only when post-relay code fails at the template level (I/O format, headers, integer width), and is prompted to touch boilerplate rather than logic; a deterministic sanitization pass (strip code fences, ensure a compilable shell) runs on every candidate regardless. Over the reported MARS runs it produced a substantive edit in a single task ($\approx 0.2\%$ of task-runs), and in $\approx 1.2\%$ of task-runs in the ablation without RAG, which indicates that Gemma 4 already emits compilable input/output wiring in most cases. Its edits are confined to I/O and compilation fixes or removal of a non-compiling fragment. It does not mask errors: the only failing task it edited stayed failing, and the task it helped passed through a legitimate output-format correction.

\section{Baseline Selection}
\label{app:baselines}

Our comparisons use the full $165$-task split and the same final ExecEval evaluation within each language block. The generator backbone is fixed within each table block, while decoding and intermediate execution follow the logged requirements of each model and method. CodeSIM and PairCoder retain their method flows but use adapters for our dataset, model endpoint, and execution service. Published numbers for the systems below use different backbones, splits, and in several cases a different task formulation, so they cannot be transferred; each would have to be ported and rerun under our protocol. PairCoder was the one additional system whose Python-specific core could be integrated without redesigning its Navigator/Driver method, and we report it in Table~\ref{tab:backbones}. For the rest we record the concrete obstacle.

\textbf{LDB} \citep{zhong2024ldb} is a debugger rather than an end-to-end generator: it presupposes a candidate program from an external generator. Its released pipeline builds control-flow graphs from the Python AST, segments Python programs into basic blocks, and records Python runtime variables with a custom tracer; the TransCoder setting uses C++ only as the \emph{source} language, while the program being debugged remains Python. Supporting C++17 would require a new control-flow/basic-block tracer and runtime-state collector.

\textbf{LPW} \citep{lei2025lpw} is a single-model plan--verify--refine workflow rather than a multi-agent system. Although its repository contains CodeContests data, the implementation is built on \texttt{PyGenerator}/\texttt{PyExecutor} and asks the model to insert Python \texttt{print} statements at individual lines, comparing runtime values against an LLM-generated plan verification. A faithful C++17 port needs compiler-safe source instrumentation, trace parsing, and redesigned extraction and repair prompts, and its two iterative phases allow up to twelve iterations each.

\textbf{MapCoder} \citep{islam2024mapcoder} is the earlier framework from the CodeSIM authors and is outperformed by CodeSIM in comparable GPT-based settings; we therefore rerun the stronger successor.

\textbf{PairCoder} \citep{zhang2024paircoder} \emph{is} reported, in Python. We ran the official Navigator/Driver implementation at upstream commit \texttt{ac7ce88} through an adapter to our task loader, model endpoint, and ExecEval sandbox, with Gemma 4 as the generator and a $4096$-token completion budget. Its upstream plan request uses temperature $0.8$ and draws five completions in one call; the remaining generation stages use their released settings. The plan stage clusters these candidates with \texttt{text-embedding-3-large}, the one component we could not replace with an open-weights model. Embedding traffic is included in the token and call counts of Table~\ref{tab:backbones}. Around $11\%$ of the $165$ tasks ended in a truncated completion and are counted as failures, making the completion budget a material limitation of the reported $0.705$. We have no C++17 number for it: its upstream prompts, code extraction, and lint logic are Python-specific.

\textbf{MaintainCoder} \citep{wang2025maintaincoder} targets maintainability under changing requirements, not one-shot functional correctness. Its protocol applies requirement modifications after an initial solution and reports Pass@5, AST similarity, change volume, maintainability index, and cyclomatic complexity. Reproducing it here would mean constructing a C++17 dynamic benchmark with requirement changes and hidden tests, not running a baseline.

\textbf{Xolver} \citep{hosain2025xolver} targets cross-problem experience accumulation with episodic and shared memory, a planner, dynamic agents, a judge, and a final verifier, at a substantially higher inference budget; its LiveCodeBench results are averaged over $32$ inference runs, which for $165$ tasks would be $5{,}280$ multi-agent executions per model. The authors describe the released code as preliminary, with hard-coded paths and a default of five examples and two agents against the paper's three.

Beyond the ports, none of these systems was published with our backbones, so each additionally requires provider adaptation, response-parsing validation, and token-budget alignment. As a scale reference, our own CodeSIM rerun averages ${\approx}817$\,s per task.

\end{document}